\documentclass[
 reprint,
 pre,
 superscriptaddress,
]{revtex4-2}

\usepackage{amsmath}
\usepackage{graphicx}
\usepackage{xcolor}

\begin{document}


\title{Edge of chaos as a guiding principle for modern neural network training}


\author{Lin Zhang}
\affiliation{Department of Physics, National University of Singapore, Singapore 117551}
\author{Ling Feng}
\affiliation{Department of Physics, National University of Singapore, Singapore 117551}
\affiliation{Institute of High Performance Computing, A*STAR, Singapore 138632}
\author{Kan Chen}
\affiliation{Department of Mathematics, National University of Singapore, Singapore 119076}
\affiliation{Risk Management Institute, National University of Singapore, Singapore 119613}
\author{Choy Heng Lai}
\affiliation{Department of Physics, National University of Singapore, Singapore 117551}


\date{\today}

\begin{abstract}
The success of deep neural networks in real-world problems has prompted many attempts to explain their training dynamics and generalization performance, but more guiding principles for the training of neural networks are still needed. Motivated by the edge of chaos principle behind the optimal performance of neural networks, we study the role of various hyperparameters in modern neural network training algorithms in terms of the order-chaos phase diagram. In particular, we study a fully analytical feedforward neural network trained on the widely adopted Fashion-MNIST dataset, and study the dynamics associated with the hyperparameters in back-propagation during the training process.
We find that for the basic algorithm of stochastic gradient descent with momentum, in the range around the commonly used hyperparameter values, clear scaling relations are present with respect to the training time (epochs) during the ordered phase in the phase diagram, and the model's optimal generalization power at the edge of chaos is similar across different training parameter combinations. In the chaotic phase, the same scaling no longer exists. The scaling allows us to choose the training parameters to achieve faster training without sacrificing performance.
In addition, we find that the commonly used model regularization method - weight decay - effectively pushes the model towards the ordered phase to achieve better performance. Leveraging on this fact and the scaling relations in the other hyperparameters, we derived a principled guideline for hyperparameter determination, such that the model can achieve optimal performance by saturating it at the edge of chaos. Demonstrated on this simple neural network model and training algorithm, our work improves the understanding of neural network training dynamics, and can potentially be extended to guiding principles of more complex model architectures and algorithms.
\end{abstract}


\maketitle

\section{Introduction}

In the last decade, deep neural networks have achieved remarkable success in many areas~\cite{lecun2015deep}, including visual object recognition~\cite{krizhevsky2012imagenet}, speech recognition~\cite{hinton2012deep}, natural language processing~\cite{collobert2011natural}, and reinforcement learning \cite{silver2016mastering}. This great practical success of deep neural networks has attracted many researchers to build frameworks to interpret different aspects of deep neural networks~\cite{charles2018interpreting}, such as applying theories in glassy systems to understand the loss landscape and the dynamics within it~\cite{choromanska2015loss,baity2018comparing,geiger2019jamming}, using information bottleneck theory to interpret the training dynamics~\cite{shwartz2017opening}, utilizing random matrix theory to study the generalization performance~\cite{mahoney2019traditional,advani2020high}. These frameworks contributed towards various properties of the learning mechanism and the generalization of deep neural networks. However, the clear understanding of them is still active ongoing research, including tuning training parameters and choosing regularization strength.

Motivated by the idea that maximized computational capability emerges at the edge of chaos~\cite{beggs2003neuronal, langton1990computation,packard1988adaptation}, many different models of neural activity and brain networks have been studied as dynamical systems, and their ordered/chaotic behavior and computational properties have been investigated. In a seminal paper~\cite{sompolinsky1988chaos}, \citeauthor{sompolinsky1988chaos} studied the chaotic dynamics of a network of randomly connected neurons with the dynamic mean-field theory (DMFT)~\cite{sompolinsky1981dynamic,sompolinsky1982relaxational,derrida1987dynamical}. Then with extended DMFT, the fixed points to chaos transition was found in random neural networks with more realistic architectures and firing dynamics~\cite{kadmon2015transition}. The ``reservoir computing" model~\cite{lukovsevivcius2012reservoir} utilizes this kind of random neural network to do different practical tasks; therefore, its computational properties can be investigated. It has been shown that close to the edge of chaos, the computational capabilities of reservoir computing are optimal both for online computation~\cite{bertschinger2004real} and off-line computation~\cite{legenstein2007edge}. Later, it was found that the information storage and transfer of reservoir computing are maximized near the edge of chaos~\cite{boedecker2012information,haruna2019optimal}, providing an explanation for the optimal computational capabilities.

While most of the earlier studies focused on recurrent neural networks like the reservoir computing model, many successful networks are feedforward networks. Deep feedforward networks' chaotic properties have also been studied~\cite{poole2016exponential}, in which each layer's weights are normally distributed and have an identical number of neurons, and exponential expressivity is found in such networks. More recently, Ref.~\cite{feng2019optimal} analyzed more generic feedforward networks including the state-of-the-art architectures, and showed that the whole deep neural network's asymptotic behavior is associated with the model performance. In particular, various feedforward networks' performances are optimal when their asymptotic stabilities are at the edge of chaos, even though the model's input-output computation does not undergo a recurrent process. This phenomenon was explained by the highest information transmission near the edge of chaos for generic high dimensional operators, similar to the special case of reservoir computing~\cite{boedecker2012information, haruna2019optimal}. However, how the model training algorithms bring the model to the edge of chaos is an open question.

In this paper, we explore a typical training process (stochastic gradient descent with momentum) of a simple feedforward neural network model, and study the dynamics of each hyperparameter in terms of the order-chaos phase diagram. For simplicity and analytical tractability, we study the simple network model with only one hidden (fully connected) layer which has the same number of neurons as the input layer. The data we use to train the network is one of the most popular image classification tasks called Fashion-MNIST. We find that the basic training algorithm `stochastic gradient descent with momentum' pushes the model towards the chaotic phase after initialization in the ordered phase, and the various parameters controlling the training process exhibit a scaling relation with the training duration (defined as the number of epochs during training) inside the ordered phase. Interestingly, the model performance is hardly affected by the hyperparameters in this scaling relation, suggesting that one can accelerate the training without sacrificing performance. Based on this scaling relation, we show that the model can be controlled to stay at the edge of chaos through specific regularization strength, and it will consequently achieve the best generalization performance compared with other parameter sets as predicted by the edge of chaos theory. We also provide a method to approximate this specific regularization strength.

\section{Edge of chaos for artificial neural networks}
For generic discrete dynamical systems $\pmb x_{t+1}=\pmb f(\pmb x_t)$, where $\pmb x_t$ is a vector of $N$ dimensions, reference~\cite{feng2019optimal} shows that at the edge of chaos condition:
\begin{align}
\frac{1}{\sqrt{N}} \|\pmb J^*\| =1,
\label{Eqn:critical}
\end{align}
the systems passes the highest amount of information from $\pmb x_t$ to $\pmb x_{t+1}$. Here $\pmb J^*$ is the Jacobian matrix of the function $\pmb f$ evaluated at the mean value of the system's asymptotic output value $\pmb x^*=\pmb x_\infty$, and $\|\pmb J^*\|$ is the Frobenius norm of $\pmb J^*$. We will denote the left-hand side of Eq.~\ref{Eqn:critical} as Jacobian norm in the rest of the paper. 

\begin{figure}[h]
  \centering
  \includegraphics[width=8.6cm]{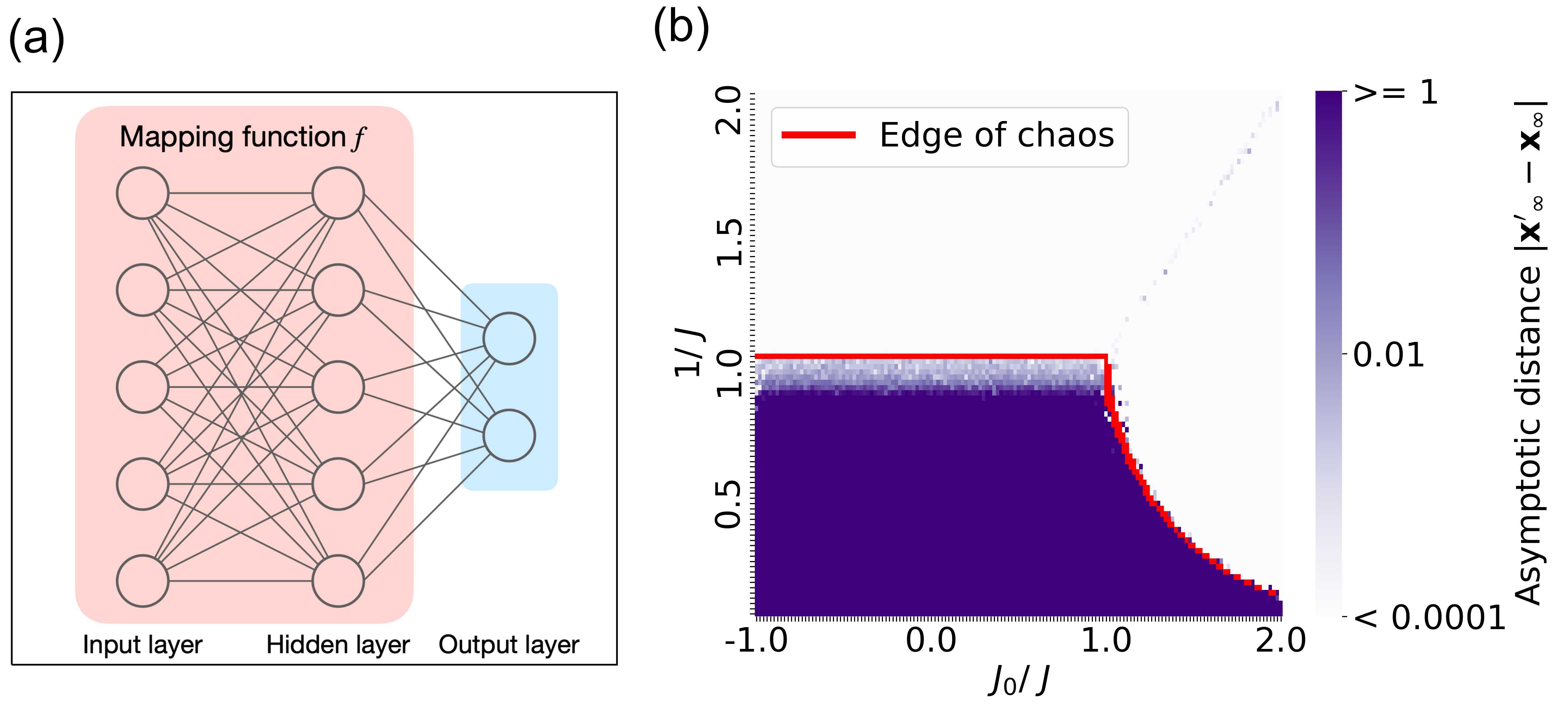}
  \caption{(a) A feedforward neural network with one hidden (fully connected) layer having the same number of neurons as the input dimension (the activation is $\tanh$). The non-linear operation in the hidden layer is extracted as the dynamical operator. The model's output layer is the widely used softmax layer and has the dimension defined by the number of classes in the specific classification task, which is Fashion-MNIST in our case. (b) Order-chaos phase diagram for this dynamical operator. The heat map is the result of asymptotic distances between two trajectories starting from two nearby inputs $\pmb x_0$ and $\pmb x'_0$. We perturb the input \(\pmb x_0\) by a Gaussian noise of mean 0 and standard deviation 0.0001, i.e., \(\pmb x_0' = \pmb x_0 + \pmb \eta\), and represent the distance between \(\pmb x_t \) and \(\pmb x'_t \) as the color of the pixel when the distance stabilizes. If the distance \(\| \pmb x_t - \pmb x'_t \|\) diminishes, the network is in the ordered phase; otherwise, it is chaotic. Since the asymptotic state is a global property independent of the initial value of \(\pmb x_0\) (Appendix~\ref{appendix:global}), we generate \(\pmb x_0\) randomly with a standard normal distribution. There are two distinct phases: the asymptotic distance converges towards 0 (white area) and diverges to a fixed size (colored area). The further away from the ordered phase, the larger is the asymptotic difference.}
  \label{Fig: phase_diagram}
\end{figure}

While reference~\cite{feng2019optimal} applied this transition boundary to different deep neural networks to demonstrate their optimal performance at the asymptotic edge of chaos, the models' complex architecture prevents an analytical solution to the edge of chaos condition. Hence, here we use a simple feedforward neural network architecture whose analytical solution is known. Specifically, our feedforward neural network has just one hidden layer (illustrated in Fig.~\ref{Fig: phase_diagram}(a)). This hidden layer has the same input and output dimensions, and uses $\tanh$ as the non-linear activation function. Mathematically, the hidden layer operator $f_{\pmb W}$ is defined as:
\begin{align}
\pmb x_\text {hidden} = f_{\pmb W} (\pmb x_\text{input})=\tanh(\pmb W \cdot \pmb x_\text{input}),
\label{Eqn:hidden}
\end{align}
where $\pmb W$ is the weight matrix. Since the weights in \(\pmb W\) are not identical in artificial neural networks, we use \(J_0/N\) to represent their mean and \(J^2/N\) to represent their variance. Therefore one can write down the matrix element $W_{ij}$ as: $W_{ij}=\frac{J_0}{N}+\frac{J}{\sqrt{N}}z_{ij}$, where $z_{ij}$ can be treated as a stochastic variable with mean 0 and variance 1. In the thermodynamic limit of \(N \rightarrow \infty\), one then retrieves the condition for the edge of chaos identical to the spin-glass boundary in the Sherrington-Kirkpatrick spin-glass model~\cite{sherrington1975solvable, de1978stability} (mathematical details in Appendix A): 
\begin{align}
J^{2} \int D_{z} \operatorname{sech}^{4}\left(J_{0} \mu+J \sqrt{q_{0}} z\right)=1,
\label{Eqn:boundary_single_integral}
\end{align}
where $\mu=\frac{1}{N} \sum_i \mu_i$, $q_{0}=\frac{1}{N} \sum_{i} \mu_{i}^{2}$, $\mu_{i}$ is the mean value of $x^*_i$, $D_{z}=\frac{1}{\sqrt{2 \pi}} e^{-\frac{z^{2}}{2}} d z$. We plot this analytical chaotic boundary (red line) in Fig.~\ref{Fig: phase_diagram}(b), which is the same as the boundary between spin-glass phase and paramagnetic/ferromagnetic phases of the Sherrington and Kirkpatrick model~\cite{sherrington1975solvable, de1978stability}. As shown in Fig.~\ref{Fig: phase_diagram}(b), the result of numerical asymptotic state (purple heatmap) matches well with this theoretical chaotic boundary. Furthermore, from Eq.~\ref{Eqn:boundary_single_integral}, we see that the asymptotic state does not depend on the inputs, meaning it is a global property depending only on the weights. We validated this global property empirically in Appendix~\ref{appendix:global}. This means our asymptotic state study is different from the studies that focus on the input-output Jacobian \cite{peng2008new,sokolic2017robust,oymak2019generalization}, which is a local property dependent on the inputs.

For generic network structures, direct assessment of the asymptotic order/chaos properties can be hard without numerical calculations. However, for this particular structure of single hidden layer neural network trained on real data, one can easily obtain the values of \(J_{0}\) and \(J\) by calculating the mean and variance of the hidden layer weights, and locate the corresponding position on the axis of Fig.~\ref{Fig: phase_diagram}(b). Therefore, it is possible to evaluate the ordered/chaotic phases of such a system purely based on its weights without going through the expensive iterative computations to evaluate the Jacobian norm in Eq.~\ref{Eqn:critical}. Since chaos in dynamical systems is an asymptotic concept, for the rest of the paper we use `edge of chaos' to refer to `asymptotic edge of chaos' for simplicity.

\section{Order to chaos transition and relation of training parameters in the ordered phase}

With the analytical framework that can map the weight distribution of the hidden layer to the order-chaos phase diagram, we can then study the back-propagation process during the model training process in the phase diagram.
In our experiment, we use the standard Fashion-MNIST image dataset~\cite{xiao2017fashion} for training and testing. This dataset contains 70,000 greyscale images of 10 types of clothing and accessories, of which 60,000 images are used for training and 10,000 images are for testing. Since the network architecture of our model requires a vector as input, we flatten the 2D images into a 1D vector.
As each sample image in the Fashion-MNIST dataset has $28 \times 28$ = 784 pixels, we design the hidden layer with 784 neurons, such that there are $784\times 784$ weights in the hidden layer. The activation function of the hidden layer is \(\tanh\) in line with the theoretical framework. We use the basic stochastic gradient descent (SGD) with momentum as the optimization algorithm. After every epoch during the training process, we calculate the mean and variance of the weights in the hidden layer to identify the ordered/chaotic phase of the hidden layer.

\begin{figure}[h]
  \centering
  \includegraphics[width=8.6cm]{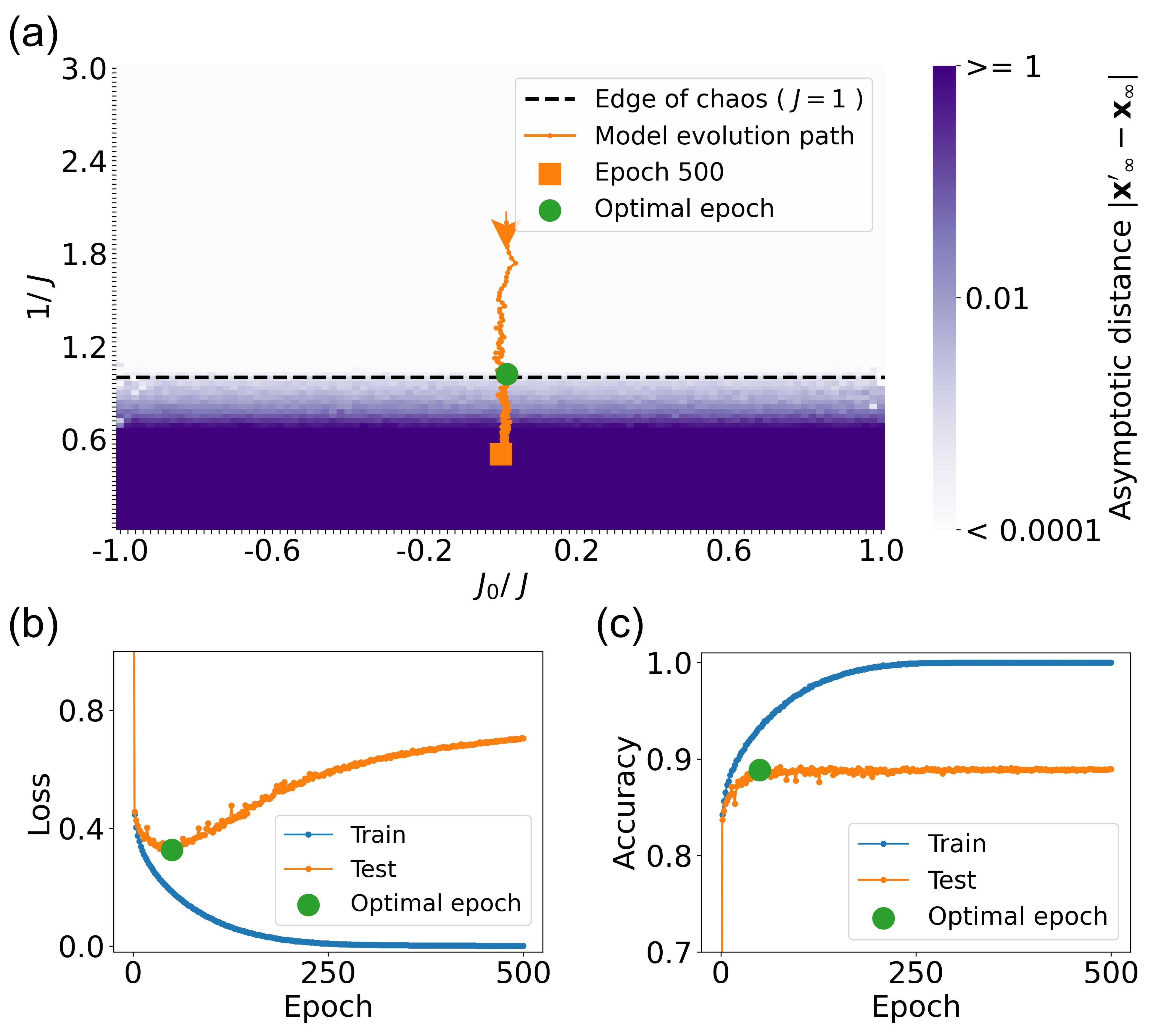}
  \caption{Model evolution path in the order-chaos phase diagram and related learning curves. The optimal epoch in (a)-(c) is the epoch with the minimum test loss. (c) shows that the optimal epoch also has the almost largest test accuracy. As the value of \(J_0\) is very small during training, we only plot the order-chaos phase diagram for \(-1<{J_0}/{J}<1\).}
  \label{Fig:model_evolution}
\end{figure}

In the training process of a neural network with back-propagation, the training data are repeated fed into the model to optimize the weights iteratively. Each time the whole set of training sample is used for one round of back-propagation, the training time is defined as incremented by one {\it epoch}. Hence, at the beginning of each epoch, we calculate $J_0$ and $J$ from the mean and variance of the hidden layer's weights, and map its coordinates to the order-chaos phase diagram for the first 500 epochs. 
The trajectory in Fig.~\ref{Fig:model_evolution}(a) represents the model's evolution path during training. We initialize the hidden layer weight matrix \(\pmb W\) with \(J_0=0\) and \(J=0.5\), such that the hidden layer starts from the ordered phase, i.e., the white region in the figure. As the training proceeds, the model evolves from the ordered phase towards the chaotic phase. When the model is around the edge of chaos $(J=1)$, the test loss is the lowest as Fig.~\ref{Fig:model_evolution}(a)-(b) shows, indicating optimal model performance around this point. This demonstrates that the model optimality near the edge of chaos also holds for single hidden layer neural networks, even though this network is a feedforward process rather than an iterative recurrent computation.

\begin{figure}[h]
  \centering
  \includegraphics[width=8.6cm]{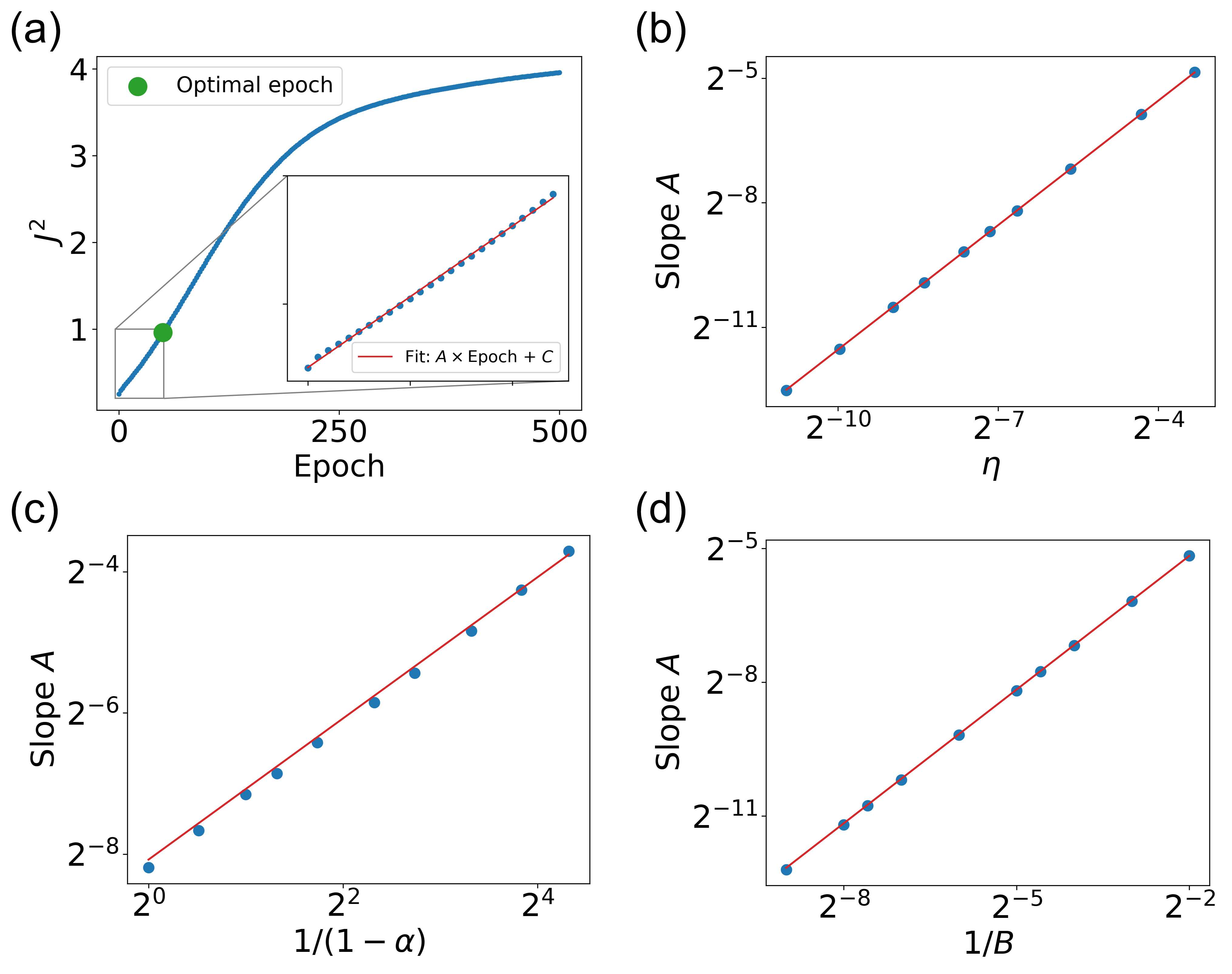}
  \caption{Scaling between the weight variance and the training parameters. Blue dots are empirical observations and the red lines are linear regression lines. (a) A typical evolution path of the variance \(J^2\). The data is fitted with $J^2 = A \times \text{epoch} + C$ when \(J^2<\)1, i.e., when the model is inside the ordered phase. (b) The linear fit of $A$ vs. $\eta$ for fixed $\alpha=0$ and $B=32$ ($\eta \in[5e-4,1e-1]$). (c) The linear fit of $A$ vs. $1/(1-\alpha)$ for fixed $\eta=0.01$ and $B=32$ ($\alpha \in[0,0.95]$). (d) The linear fit of $A$ vs. $1/B$ for fixed $\eta=0.01$ and $\alpha=0$ ($B \in[4,512]$).}
  \label{Fig:J2}
\end{figure}

As \(J_0\) is very small during training (see Fig.~\ref{Fig:model_evolution}(a)), we can deduce the order/chaos states of the model from the value of \(J\) alone. Fig.~\ref{Fig:J2}(a) shows the value of \(J^2\) over time. In the ordered phase (\(J^2<\)1), \(J^2\) will increase linearly with time, 
\begin{align}
J^2 = A \times \text{epoch} + C,
\label{Eqn:J2}
\end{align}
where $A$ depends on training parameters and $C$ depends on the initial weights of the model. The slope $A$ influences how fast a model arrives at the edge of chaos (\(J^2=\)1), which is the optimal point as shown in Fig.~\ref{Fig:model_evolution}(a). Furthermore, we find that the relation between slope $A$ and training parameters is:
\begin{align}
A = \frac {\eta}{(1-\alpha)B} \cdot D,
\label{Eqn:slope}
\end{align}
where \(\eta\) is the learning rate, \(\alpha\) is the momentum of SGD, \(B\) is the mini-batch size, and $D$ is a constant which depends on network structure and training data. 
To elaborate on the parameters, the SGD with momentum training algorithm randomly selects $B$ number of samples to compute the back-propagation gradient each time, while \(\eta\) determines how much to adjust the weight based on the gradient calculated, and \(\alpha\) determines how much previous gradients to incorporate in the current gradient computation.
This linear relation in the ordered phase is further illustrated in Fig.~\ref{Fig:J2}(b)-(d). Such linear scaling is indicative of the simplicity of the dynamics inside the ordered phase as compared to the chaotic phase. It is also interesting that the stochasticity from using mini-batches is not a dominant force in driving the model towards the minimal loss function; otherwise, it would be a random walk process with $J^2$ scaling with $\eta^2$ instead of $\eta$.

\begin{figure}[h]
  \centering
  \includegraphics[width=8.6cm]{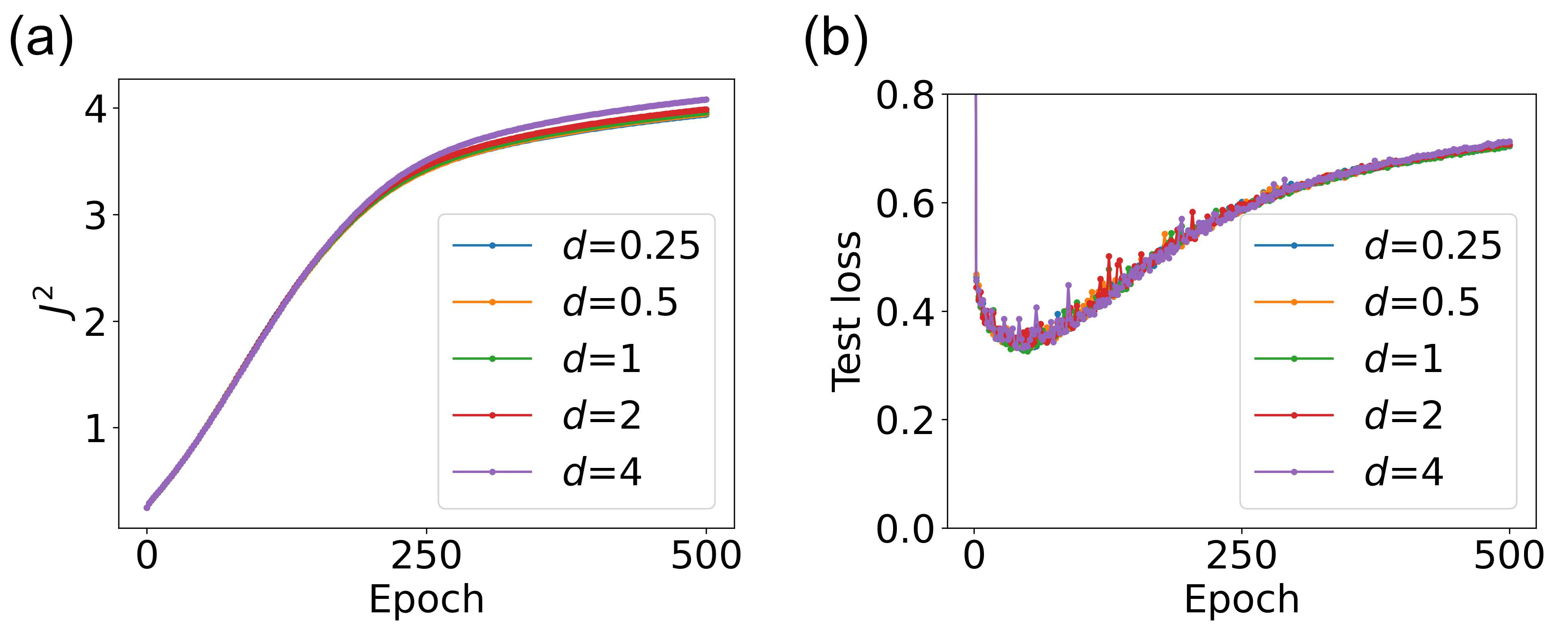}
  \caption{Model evolution with the same hyperparameter scale factor. (a) Variance of the model's weights \(J^2\) increases with similar rate vs. training epochs. The baseline \(d=1\) corresponds to training parameter combination: $\eta=0.02, \alpha=0.5$ and $B=32$. We choose this combination to save model training time. To get the same scale factor but different parameter combinations, we fix  $\alpha$ and multiply $\eta$ and $B$ by the same coefficient $d$. (b) Test losses for different combinations of training parameters with the same scale factor.}
  \label{Fig:J2_same_scale_factor}
\end{figure}

Now we denote \({\eta}/{(1-\alpha)B}\) as the scale factor. This scale factor \({\eta}/{(1-\alpha)B}\) is similar to the noise scale \(\eta S/{(1-\alpha)B}\) in the stochastic differential equation (SDE) approximation of SGD~\cite{smith2018bayesian}, where $S$ is the training sample size. It was argued in Ref.~\cite{smith2018bayesian} that for training algorithms with the same noise scale, the models would evolve similarly as their SDE approximation are the same. This is indeed found in our results as demonstrated in Fig.~\ref{Fig:J2_same_scale_factor}. In Fig.~\ref{Fig:J2_same_scale_factor}(a), for different hyperparameter combinations but with the same scale factor, the evolutions of \(J^2\) are almost the same. Further in Fig.~\ref{Fig:J2_same_scale_factor}(b), generalization performances are also shown to be similar.
However, our findings further distinguish the two different phases of such relations, that the model evolutions behave with distinctive patterns in the ordered (linear) and chaotic phases (sub-linear). The linear scaling in the ordered phase is likely due to the lack (or low amount) of local minima that traps the gradient descent process, yet the sub-linear scaling in the chaotic phase is due to the exact opposite. The deeper in the chaotic phase (larger $J^2$ value), the more sub-linear is the relation.

\begin{figure}[h]
  \centering
  \includegraphics[width=8.6cm]{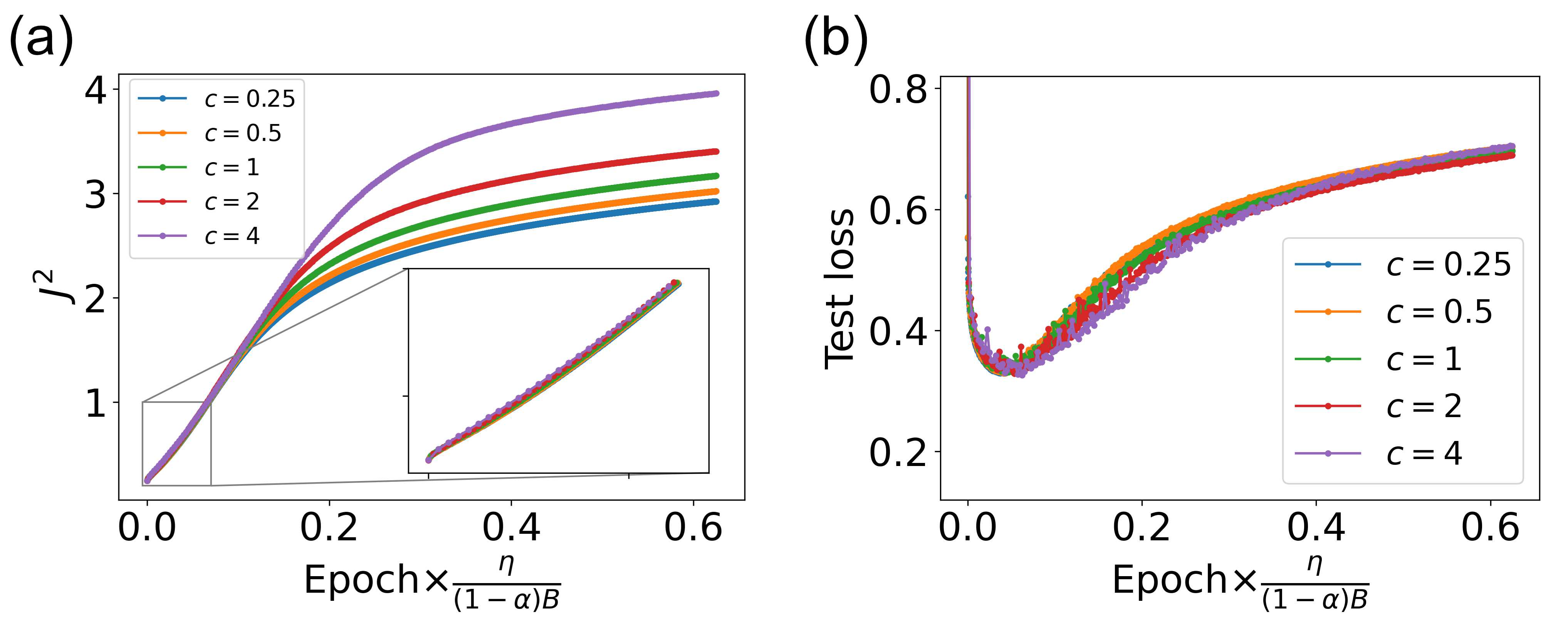}
  \caption{Model evolution with different hyperparameter scale factors. (a) Variance of the model's weights \(J^2\) for different combinations of training parameters with scale factor given by \(0.0003125 * c\). The value \(0.0003125\) is the scale factor corresponding to the default training parameters ($\eta=0.01, \alpha=0$ and $B=32$). For ease of comparison, we rescale the epoch by \({\eta}/{(1-\alpha)B}\). (b) Test losses for different combinations of training parameters with scale factor given by \(0.0003125 * c\).}
  \label{Fig:J2_rescale}
\end{figure}

Interestingly, for different scale factors, the Eq.~\ref{Eqn:J2} and Eq.~\ref{Eqn:slope} indicate that the training processes in the ordered phase are also similar if the training epochs are scaled by this scale factor. The only difference is the evolution rate of \(J^2\). In Fig.~\ref{Fig:J2_rescale}(a), with rescaled epoch coordinate, we find that the evolution of \(J^2\) becomes very similar in the ordered phase (inset graph). In the chaotic phase, however, such linear scaling disappears, likely due to the more complex loss landscape.
In addition, the models' generalization performance is also similar with such scaling in epochs, which is shown in Fig.~\ref{Fig:J2_rescale}(b). This phenomenon suggests that in certain parameters ranges around the default value (set in deep learning libraries like TensorFlow) of scale factor, the models have similar generalization performances. Meanwhile, large scale factor will help the model arrive at the edge of chaos faster (with fewer epochs). Therefore, when tuning the training parameters, one needs to pay more attention to this scale factor which is the combined effect of the individual parameters. 

Since the model's performance is optimal at the boundary between order and chaos and there is clear linear scaling in the ordered phase, in practice, one can focus on the model dynamics in the ordered phase during the training process while avoiding the chaotic phase. This is because in the chaotic phase, the model's stability evolution is sensitive to the values of scale factor in a sub-linear manner that is harder to control as shown in Fig.~\ref{Fig:J2_rescale}(a). This sub-linear relation indicates that the noise scale \(\eta S/{(1-\alpha)B}\) in the training dynamics plays a more important role when the model evolves to the chaotic phase.

\section{Principled regularization to the edge of chaos}

As we have seen, the ordered phase of the model training process can be more easily leveraged to control the training, and the optimal model weights occur at the edge of chaos. Hence, it would be ideal if we can control the training process such that the model stays at the edge of chaos and explore optimal weights configurations. This is exactly what typical model regularization techniques can do~\cite{goodfellow2016deep}. While regularizations are used to improve the model's generalization power, over-regularization may also degrade it. Although there are many efforts to understand the mechanism of regularization methods~\cite{krogh1992simple,zhang2018three,golatkar2019time}, how to predict optimal regularization strength is still an open problem. In practice, the optimal regularization strength is usually found by random search~\cite{bergstra2012random} which is computationally expensive. 
The edge of chaos theory then leads to the definition of the optimal regularization strength: it is the one that can bring the model to the edge of chaos and stay there. In other words, when the various training parameters act together with the regularization strength to bring the model's steady state at the edge of chaos, its performance should be the most optimal.

\begin{figure}[h]
  \centering
  \includegraphics[width=8.6cm]{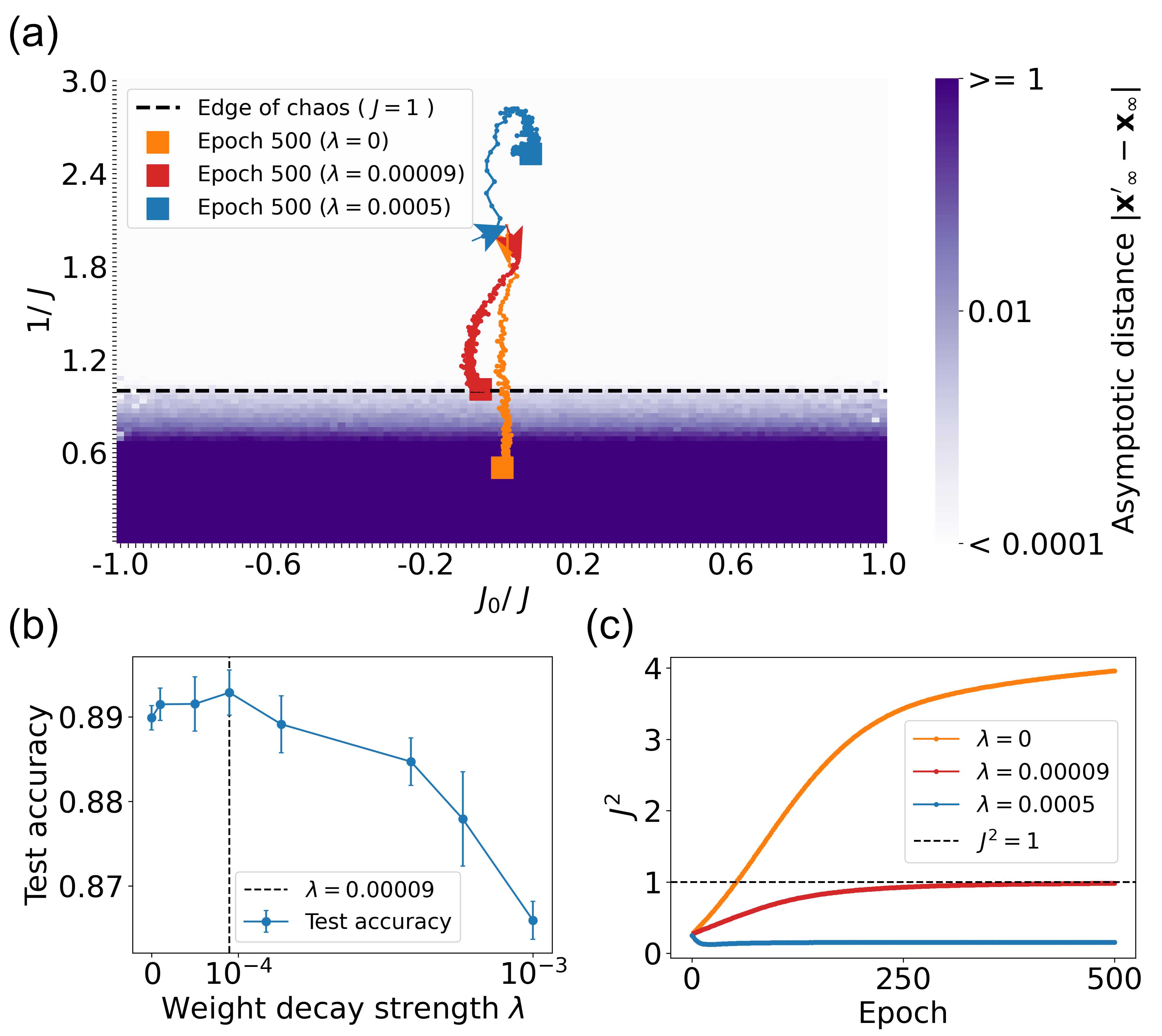}
  \caption{Effect of weight decay on the model's order/chaos properties. (a) Model evolution paths for three typical weight decay strengths $\lambda = 0, 9\times 10^{-5}, 5\times 10^{-4}$. The dashed line represents the edge of chaos. (b) Test accuracy vs. weight decay strength \(\lambda\). The error bars represent the standard deviation over 10 experiments. The best test accuracy happens when the model saturates at the edge of chaos. (c) Evolution of \(J^2\) for three typical weight decay strengths. The horizontal dashed line refers to the edge of chaos.}
  \label{Fig:model_evolution_wd}
\end{figure}

To test this hypothesis, we apply one of the most common types of regularizations: weight decay to the same model used in Fig.~\ref{Fig:model_evolution}. This method penalizes the size of the weights in the model by a fixed fraction denoted by $\lambda$. 
As Fig.~\ref{Fig:model_evolution_wd}(a) shows, different weight decay strengths \(\lambda\) can bring the model to different phases. When \(\lambda=0.00009\), the model ends right at the edge of chaos; meanwhile, its test accuracy happens to be the largest as shown in Fig.~\ref{Fig:model_evolution_wd}(b), indicating optimal model generalization power. Thus, the regularization strength maintaining the model at the edge of chaos is the optimal strength, as we suggested earlier. From the view of underfitting and overfitting, when \(\lambda<0.00009\), the model ends at chaotic phase and suffers from overfitting; when \(\lambda>0.00009\), the model ends at ordered phase and underfits. 
Therefore, regularizing the model to the edge of chaos will let the model achieve a balance between underfitting and overfitting. This phenomenon further justifies the importance of the edge of chaos in artificial neural networks' training process, and it is also another evidence about maximized computational capability at the edge of chaos~\cite{langton1990computation,packard1988adaptation}, even for a feedforward process. 

Next, we will show how to calculate this optimal choice of \(\lambda\) from the principle of the edge of chaos and the linear scaling relations of the training parameters. We then further prove that this value is actually independent of training parameters.
To maintain the model at the edge of chaos \((J^*=1)\), we need to regularize the training such that \(J^2\) saturates at 1 (as the red line in Fig.~\ref{Fig:model_evolution_wd}(c)). Here we use \(t\) to represent the number of mini-batches having been trained, such that \(\pmb w_t\) refers to the values of the weight matrix in the hidden layer after \(t\) mini-batches of training since the beginning (model initialization). The equation that describes the weight updates of SGD with momentum is then given by:
\begin{align}
\Delta \boldsymbol{w}_{t}=\boldsymbol{v}_{t}=\alpha \boldsymbol{v}_{t-1}-\eta\left(\boldsymbol{g}_{t}+2 \lambda \boldsymbol{w}_{t}\right),
\label{weight_decay}
\end{align}
where $\pmb v_{t}$ is the amount of weight adjustments, and $\pmb g_{t}$ is the gradient estimate. Recursively substituting the equation for \(\boldsymbol{v}_{t}\), we arrive at the following relation (Appendix~\ref{appendix:control}):
\begin{align}
\Delta \pmb w_t \approx - \eta \sum_{m=0}^{t-1} \alpha^m \pmb g_{t-m} -2\eta \lambda \sum_{m=0}^{t-1} \alpha^m \pmb w_{t-m}.
\label{weight_decay_approximation}
\end{align}
Here the first term on the right-hand side is the weight update without weight decay, and it will linearly increase the value of \(J^{2}\) when \(J^{2} \leq 1\) as depicted in Fig.~\ref{Fig:J2}(a).
Therefore, to control the \(J^2\) to be at 1, we need to adjust the second term in Eq.~\ref{weight_decay_approximation} such that it decreases \(J^2\) by the same amount per mini-batch at the edge of chaos condition \(J^{*2}=1\). Assuming \(\Delta \pmb w_t \ll \pmb w_t\), i.e., the changes in weights are small for every mini-batch, we have the following relation (Appendix~\ref{appendix:control}):
\begin{align}
\lambda &\approx \frac{(1-\alpha) B}{\eta} \cdot \frac{A}{4 S J^{* 2}} \label{Eqn:lambda1} \\
&= \frac{D}{4 S J^{* 2}} = 0.00005, \label{Eqn:lambda2}
\end{align}
where $S$ is the training sample size. The empirical relation Eq.~\ref{Eqn:slope} is used for the derivation from Eq.~\ref{Eqn:lambda1} to Eq.~\ref{Eqn:lambda2}. This estimated value of weight decay strength 0.00005 is close to the empirical value \(\lambda=0.00009\) needed to control the model to saturate at \(J^{*2}=1\). Moreover, because $D$ only depends on the network structure and training data, this optimal weight decay strength \(\lambda\) is independent of training parameters, i.e., learning rate, mini-batch size, and momentum.

\section{Discussion}

Edge of chaos has been long proposed as the state of healthy brain neural networks~\cite{beggs2003neuronal}, and was also studied in both recurrent~\cite{toyoizumi2011beyond} and feedforward~\cite{feng2019optimal} networks. Here we demonstrate that such principle not only explains the optimality of neural networks, but also can be used to achieve such optimality by controlling the model's dynamical evolution in the order-chaos phase diagram during training. Using the data from a popular deep learning benchmark, we adopt a simple structure that maps to the theoretical spin-glass  models~\cite{little1974existence,hopfield1982neural,amit1985spin} to study the order-chaos phase transition during the modern neural network training process based on back-propagation. The training process with SGD with momentum is found to move the model from order to chaos, and has the highest generalization performance at the transition between these two phases.

Training hyperparameters in modern algorithms usually require a lot of experimentation, as there is a need for more theoretical guidance on the optimal weights to be reached through gradient descent. With the help of the order-chaos phase diagram, we see that the model's weights evolve very differently, allowing us to leverage the linear scaling property inside the ordered phase to define the training hyperparameters in a very principled manner. Two principles helped with the training: the linear scaling of the training parameters allow us to set training parameters such that optimal weights can be reached faster; the optimal regularization strength can be derived based on the principle that the model stays at the edge of chaos indefinitely while exploring more optimal weights.

While our work is demonstrated out on an analytical model mapped to the spin-glass model, some open questions remain. The linear and sub-linear scaling in the ordered and chaotic phase point to some deeper connection with the loss landscape in these two different phases, which are not usually considered in the investigations using stochastic differential equations~\cite{mandt2015continuous,li2017stochastic,smith2018bayesian}. Further studies could improve a more refined understanding of the training process.
Another more practical but interesting direction is to investigate more generic network structures and other types of training algorithms in this framework of order-chaos phase transition, so as to generate more principled training process.

\appendix

\section{Derivation of chaotic boundary}
\label{appendix:derivation}

The generic form of the solution of the attractor in an arbitrary dynamical system can be written as:
\begin{align}
x_i^* = \mu_i + \sigma_i z_i, \quad  \forall i \in \{1,2,...,N\},
\label{xi}
\end{align}
where \(\mu_i\) represents the mean value of \(x^{*}_i\) and \(\sigma_i\) is the standard deviation of \(x^{*}_i\), such that \(z_i\) is a normalized random variable. 
The asymptotic attractor $x^*$ for the hidden layer operator Eq.~\ref{Eqn:hidden} should satisfy the relation:
\begin{align}
\pmb x^* = \tanh(\pmb W \cdot \pmb x^*).
\label{asymptotic_attractor}
\end{align}
Then Eq.~\ref{asymptotic_attractor} can be written as:
\begin{align}
x_i^* = \tanh (\sum_j W_{ij} x_j^*),
\label{xi*}
\end{align}
where \(W_{i j}=\frac{J_{0}}{N}+\frac{J}{\sqrt{N}} z_{i j}\). Then the square of the Jacobian norm will be:
\begin{align}
\frac{1}{N} \|\pmb J^*\|^2 &= \frac{1}{N} \sum_{i, j}\left(\frac{\partial \left(\tanh (\sum_j W_{ij} x_j)\right)}{\partial x_{j}}\right)^{2}\Bigg\rvert_{x_j=\mu_j} \notag \\
&= \frac{1}{N} \sum_{i, j}\left(\operatorname{sech}^{4}\left(\sum_j W_{i j} \mu_{j}\right) \cdot W_{i j}^{2}\right) \notag \\
&= \frac{1}{N} \sum_i\left(\operatorname{sech}^{4}\left(\sum_j W_{i j} \mu_{j}\right) \cdot \sum_j W_{i j}^{2}\right). \label{Eqn:jacobian_norm}
\end{align}
As the mean value of $\sum_j W_{i j}^{2}$ is \(N\left(\frac{J^{2}}{N}+\frac{J_{0}^{2}}{N^{2}}\right)\), we can write Eq.~\ref{Eqn:jacobian_norm} as:
\begin{align}
\frac{1}{N} \|\pmb J^*\|^2 & \approx \left(\frac{J^{2}}{N}+\frac{J_{0}^{2}}{N^{2}}\right) \sum_i \operatorname{sech}^{4}\left(\sum_j \frac{J_{0} \mu_{j}}{N}+\frac{J \mu_{j}}{\sqrt{N}} z_{i,j}\right).
\end{align}
Thus, in the thermodynamic limit of \(N \rightarrow \infty\), the condition for the edge of chaos is:
\begin{align}
\frac{J^2}{N} \sum_{i} \operatorname{sech}^{4}\left(\sum_{j}\left(\frac{J_{0} \mu_{j}}{N}+\frac{J \mu_{j}}{\sqrt{N}} z_{i j}\right)\right)=1.
\label{Eqn:boundary_single}
\end{align}
where $\mu_{j}$ is the mean value of $x^*_j$.

By the central limit theorem, the sum over $j$ in Eq.~\ref{Eqn:boundary_single} approximately follows a normal distribution with mean value of \(\frac{J_{0}}{N} \sum_i \mu_i\) and variance of \(\frac{J^2}{N} \sum_{i} \mu_{i}^{2}\). Therefore we can replace the summation over $i$ in Eq.~\ref{Eqn:boundary_single} by an integral over Gaussian measure in Eq.~\ref{Eqn:boundary_single_integral}.

\section{Global property of asymptotic states}
\label{appendix:global}

To validate our theoretical results that the asymptotic state of a model is a global property independent of the specific input data, we test it on the models used in the main paper. After training at each epoch, we perturb each of the 10000 test images by the same random amount, i.e., each pixel is added a random noise distributed randomly with mean 0 and standard deviation 0.0001. Then we apply the same asymptotic state analysis as the one in Fig.~\ref{Fig: phase_diagram}(b) in the main paper, and examine the distributions of the asymptotic distances after each epoch.
 
\begin{figure}[h]
  \centering
  \includegraphics[width=8.6cm]{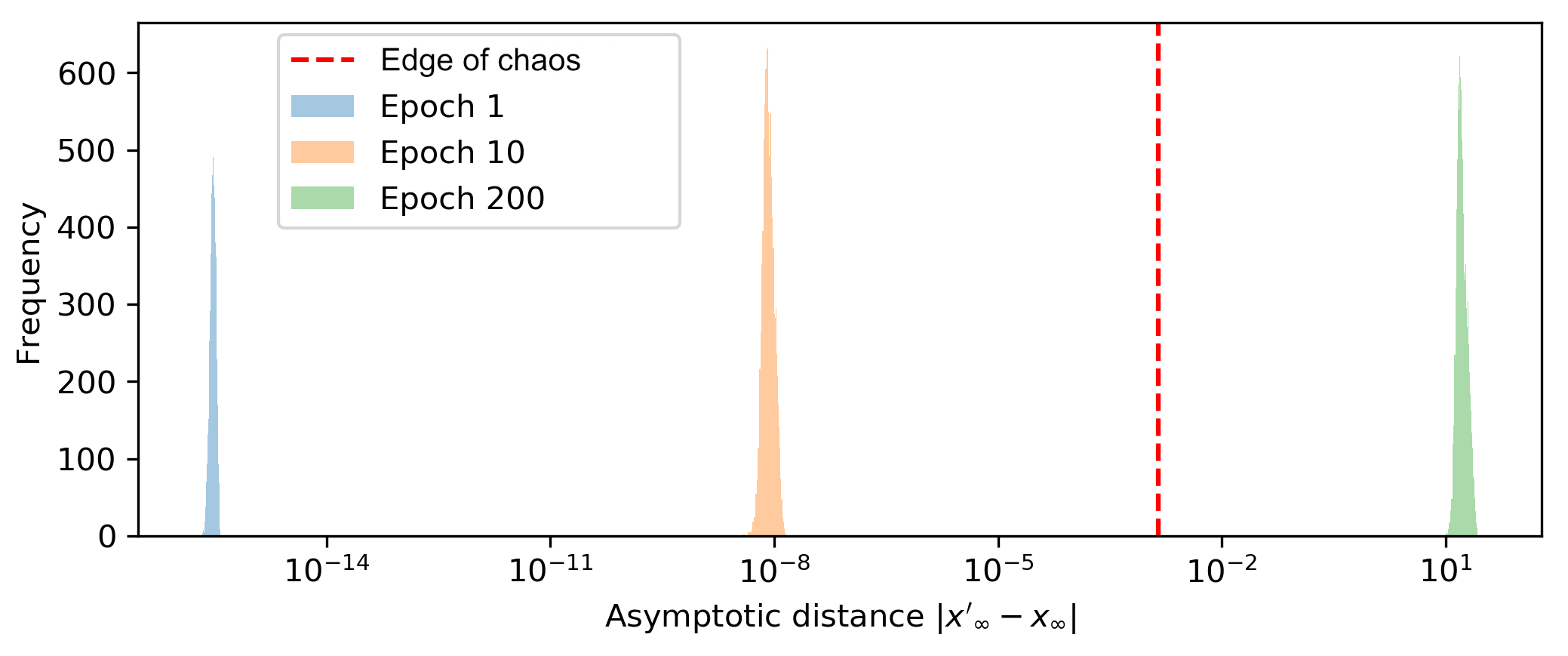}%
  \caption{Distributions of asymptotic distances across 10000 testing images. Each distribution is from an epoch during the model's training.}
  \label{global_property}
\end{figure}

Since the ordered phase corresponds to \(|x'_\infty - x_\infty| \to 0\), we use the distance at the 50th iteration. If \(|x'_\tau - x_\tau| \ll |x'_0 - x_0|\), the model is in the ordered phase. We use \(\tau=50\) due to the limit of the numerical precision of our program. In the chaotic phase, the asymptotic distances saturate around constant values, which are used as \(|x'_\infty - x_\infty|\). Fig.~\ref{global_property} shows the distributions of the asymptotic distances for the model at epoch 1, 10, and 200. It can be seen that the distributions are narrow, indicating homogeneous asymptotic behaviors across the test images used.

\section{Quantitative control on the edge of chaos through weight decay}
\label{appendix:control}

Substituting Eq.~\ref{weight_decay} in the main paper iteratively into itself yields:
\begin{align}
\pmb v_t &=  \alpha \pmb v_{t-1} - \eta (\pmb g_t + 2\lambda \pmb w_t) \notag \\
&=  \alpha (\alpha \pmb v_{t-2} - \eta (\pmb g_{t-1} + 2\lambda \pmb w_{t-1})) - \eta (\pmb g_{t} + 2\lambda \pmb w_{t}) \notag \\
&\  \ \vdots \notag \\
&=\alpha^t \pmb v_0- \eta \sum_{m=0}^{t-1} \alpha^m \pmb g_{t-m} -2\eta \lambda \sum_{m=0}^{t-1} \alpha^m \pmb w_{t-m} \notag \\
&\approx- \eta \sum_{m=0}^{t-1} \alpha^m \pmb g_{t-m} -2\eta \lambda \sum_{m=0}^{t-1} \alpha^m \pmb w_{t-m}.
\label{weight_decay_full}
\end{align}
The last step is due to \(\alpha<1\) and \(t  \gg 0\).

The first term in Eq.~\ref{weight_decay_full} is the only term in the case of no weight decay, i.e., \(\lambda=0\), and it increases \(J^2\) by approximately a fixed amount at each mini-batch, as shown in Fig.~\ref{Fig:J2}(a) and Fig.~\ref{Fig:model_evolution_wd}(c). We empirically obtain this increment (per mini-batch when \(J^2 \leq 1\) is small): 
\begin{align}
\Delta J_{+}^{2}=\frac{A}{S/B}=\frac{A \cdot B}{S},
\label{J+}
\end{align}
where $A$ is the slope in the linear relation between \(J^2\) and epoch, $B$ is the mini-batch size, $S$ is the training sample size.

The second term in Eq.~\ref{weight_decay_full} decreases \(J^2\) at each mini-batch, and at the saturating value of \(J^{*2}\) such decrement cancels out with the increment \(\Delta J^2_+\). Since the changes in weights are effectively small at each mini-batch compared with the size of the weights, we make the assumption:
\begin{align}
\pmb w_t \approx \pmb w_{t-m},
\label{decay}
\end{align}
for small \(m\). Hence, the second term in Eq.~\ref{weight_decay_full} decreases the variance of weights by a fraction of:
\begin{align}
1 - (1-2\eta \lambda \sum_{m=0}^{t-1} \alpha ^m)^2 \approx 4 \eta \lambda /(1-\alpha)
\end{align}
Since \(J^2\) is proportional to the variance of weights, it is decreased by the same fraction, leading to
\begin{align}
\Delta J_-^2 = 4\eta \lambda J^{*2}/(1-\alpha).
\label{J-}
\end{align}
Equating Eq.~\ref{J+} and \ref{J-}, we obtain the Eq.~\ref{Eqn:lambda1} in the main paper.

\bibliography{references_training}

\end{document}